\documentclass{article}

\usepackage{microtype}
\usepackage{graphicx}
\usepackage{subfigure}
\usepackage{booktabs} % for professional tables

\usepackage{hyperref}

\usepackage[accepted]{icml2025}

\usepackage{amsmath}
\usepackage{amssymb}
\usepackage{mathtools}
\usepackage{amsthm}

\usepackage[capitalize,noabbrev]{cleveref}

\theoremstyle{plain}

\theoremstyle{definition}

\theoremstyle{remark}

\usepackage[most]{tcolorbox}
\usepackage{enumitem}
\usepackage{caption}
\definecolor{boxgray}{gray}{0.25}
\definecolor{lightgray}{gray}{0.65}

\definecolor{HeaderBG}{RGB}{35,35,35}   % dark header
\definecolor{HeaderFG}{RGB}{255,255,255}
\definecolor{FrameGray}{gray}{0.25}
\usepackage[textsize=tiny]{todonotes}

\usepackage{xcolor}
\definecolor{lightgray}{gray}{0.45}
\newcommand{\worse}[1]{\textcolor{lightgray}{#1}}
\definecolor{sectiongray}{gray}{0.2}

\newcommand{\model}[1]{{\text{#1}}}
\newcommand{\cmdareasoning}{\model{Command-A-Reasoning}}
\newcommand{\deepseek}{\model{DeepSeek-R1}}
\newcommand{\cmdatranslate}{\model{Command-A-Translate}}
\newcommand{\gemini}{\model{Gemini-2.5-Flash}}
\newcommand{\claude}{\model{Claude-4-Opus}}

\icmltitlerunning{Unlocking Reasoning Capability on Machine Translation in LLMs}

\begin{document}

\twocolumn[
\icmltitle{Unlocking Reasoning Capability on Machine Translation in \\ Large Language Models}



\icmlsetsymbol{equal}{*}

\begin{icmlauthorlist}
\icmlauthor{Sara Rajaee}{equal,uva}
\icmlauthor{Sebastian Vincent}{cohere}
\icmlauthor{Alexandre Berard}{cohere}
\icmlauthor{Marzieh Fadaee}{cLabs}
\icmlauthor{Kelly Marchisio}{cohere}
\icmlauthor{Tom Kocmi}{cohere}
\end{icmlauthorlist}

\icmlaffiliation{uva}{University of Amsterdam}
\icmlaffiliation{cohere}{Cohere}
\icmlaffiliation{cLabs}{Cohere Labs}

\icmlcorrespondingauthor{Sara Rajaee}{s.rajaee@uva.nl}
\icmlcorrespondingauthor{Tom Kocmi}{kocmi@cohere.com}

\icmlkeywords{Machine Learning, Machine Translation, Reasoning, LLM, ICML}

\vskip 0.3in
]



\printAffiliationsAndNotice{\icmlEqualContribution} % otherwise use the standard text.

\begin{abstract}

Reasoning-oriented large language models (RLMs) achieve strong gains on tasks such as mathematics and coding by generating explicit intermediate reasoning. However, their impact on machine translation (MT) remains underexplored. We systematically evaluate several open- and closed-weights RLMs on the WMT24++ benchmark and find that enabling explicit reasoning consistently degrades translation quality across languages and models. Analysis reveals that MT reasoning traces are highly linear, lacking revision, self-correction and exploration of alternative translations, which limits their usefulness. Furthermore, injecting higher-quality reasoning traces from stronger models does not reliably improve weaker models' performance. To address this mismatch, we propose a structured reasoning framework tailored to translation, based on multi-step drafting, adequacy refinement, fluency improvement, and selective iterative revision. We curate a synthetic dataset of dynamic structured reasoning traces and post-train a large reasoning model on this data. Experiments show significant improvements over standard translation fine-tuning and injected generic reasoning baselines. Our findings demonstrate that reasoning must be task-structured to benefit MT.
%the quality of the final translation plays a more dominant role.

% that 
% Advancing reasoning ability has unlocked impressive new capabilities in large language models (LLMs) across reasoning demanding tasks. Yet, it remains unclear to what extent machine translation (MT) benefits from these reasoning skills. Our extensive analysis on open and closes-weights LLMs shows that having reasoning traces for MT does not provide additional performance improvements and in many cases it hurts the translation quality. Yet, techniques such as chain-of-the-thoughts do improve MT capabilities. Therefore, we propose a novel synthetic data generation pipeline to curate reasoning trajectories for supervised fine-tuning (SFT) LLMs that actually boost the model MT performance. We conduct an extensive set of experiments to find the most effective data recipes to inject reasoning knowledge into LLMs. Our supervised model achieves ... improvements on WMT24++ benchmark compared to other baselines. 
\end{abstract}

\section{Introduction}
% The recent success of large language models (LLMs) has been largely attributed to their reasoning ability, mostly being verbalized through \textit{thinking tokens} by them. \sv{citation missing}
% Models equipped with such a mechanism, called reasoning language models (RLMs), have demonstrated impressive performance gains across a range of complex, reasoning-intensive tasks, particularly in math and programming-oriented problems.
% However, this line of research has remained heavily biased toward a few tasks, leaving the role of reasoning in other fundamental generation tasks, such as machine translation (MT), comparatively underexplored.
Reasoning-oriented language models (RLMs), that is, large language models (LLMs) trained to spend more compute generating intermediate reasoning before answering, have recently delivered large performance gains on reasoning-heavy benchmarks, with especially strong improvements in mathematics and coding tasks~\cite{kimiteam2025kimik2openagentic,mistralai2025magistral,qwen3technicalreport,nvidia2025nvidianemotronnano2}.
This paradigm is increasingly viewed as a general strategy for improving model reliability, controllability, and performance via test-time compute scaling, making it a central design axis for the next generation of foundation models~\cite{wang2025agenttts,wang2025samplingefficient}.
As a result, these improvements have sparked interest in whether explicit reasoning can also help on other challenging generative tasks, including machine translation (MT).
%, a core capability for multilingual access to information, global communication, and cross-lingual knowledge transfer, where long-range consistency, domain constraints, and ambiguity resolution often benefit from structured deliberation~\cite{gemmateam2025gemma3technicalreport,nllbteam2022languageleftbehindscaling}.

At the same time, relative to the depth of work on math and programming evaluations, MT has seen fewer systematic studies of RLMs and reasoning-style prompting as a first-class design axis~\cite{liu2025newtrendsmodernmachine,li2026testtimescalingreasoningmodels}, leaving open whether reasoning-based scaling transfers to high-stakes real-world generative tasks.

This work bridges the gap by investigating whether and how reasoning capabilities contribute to machine translation performance. 
We begin by evaluating a diverse set of open- and closed-weight RLMs on the WMT24++ benchmark \citep{deutsch-etal-2025-wmt24}, comparing RLMs performance with and without using thinking tokens.
Contrary to findings on most reasoning-demanding tasks~\cite{jiang-etal-2025-large-language,sun2024corex}, and contrary to the common assumption that explicit reasoning is broadly beneficial across domains, our results show that enabling explicit reasoning does not improve translation quality. 
In fact, we even observe a slight decline in performance when RLMs generate reasoning traces for MT.

To better understand this behavior, we conduct a qualitative analysis of the generated reasoning trajectories in \cref{subsec:structural_analysis}. We find that these traces exhibit a linear structure lacking exploration of alternative translations, self-correction, and revision.
%Inspecting the structure of reasoning traces for machine translation, through a qualitative analysis, we notic that these trajectories have a linear structre without exploration, self-correction, and revision.
This contrasts with math and code problems, where effective reasoning involves hypothesis exploration, backtracking, and verification~\cite{marjanovic2026deepseekr,cuesta-ramirez-etal-2025-lrms}, behaviors that align well with search-based problem structures but appear mismatched with the incremental, constraint-driven nature of translation.
%Such a structure contrasts with the reasoning styles of math and code problems, where RLMs explore multiple possible solutions to find the correct answers.

We further examine whether reasoning traces generated by a stronger RLM can benefit a weaker model, enhancing the performance.
More specifically, in \cref{subsec:reasoning_injection}, we inject reasoning traces generated by a stronger RLM (called a \textit{injecting} model) into another RLM (called a \textit{receiving} model), allowing the receiving model to generate final translations conditioned on the injected reasoning. Surprisingly, even when the injected reasoning traces are of high quality and accompanied by strong translations, we do not observe an improvement in the receiving model's performance.
This suggests that simply increasing test-time compute or transferring generic reasoning traces is insufficient for improving translation, raising questions about the universality of current reasoning paradigms.
%These findings suggest that existing reasoning traces fail to benefit RLMs on MT.
% \sv{Using source RLM and target RLM is confusing because in MT normally these would be languages, so one could think source RLM = source language RLM}

% \tk{There is a missing argument to our hypothesis which I would add here or inside next paragraph: while reasoning capabilities doesn't improve MT, we also know that chain-of-thoughts helps, this gap highlighs that there is a room for improvements and we hypothetize that its only limitation of current reasoning styles...}

Based on these observations, we hypothesize that the limitation is not the absence of reasoning per se, but a misalignment between generic reasoning formats and the structural requirements of translation.
Our observation is furthermore reinforced by the success of Chain-of-thoughts reasoning that improves MT quality~\cite{nguyen-xu-2025-reasoning,10.1162/TACL.a.25}.

To test our hypothesis, in \cref{sec:structured-reasoning} we design a synthetic data generation pipeline that enforces a \textit{structured reasoning} style for MT. Our proposed reasoning template explicitly incorporates iterative refinements using a multi-turn translation setup \cite{briakou-etal-2024-translating} and a dynamic format that adapts to the difficulty of the given translation examples, enhancing the efficiency of reasoning length. 
We show that reasoning can be beneficial for MT when it is explicitly shaped around error revision, constraint satisfaction, and iterative refinement, rather than leaving it up on model for generic reasoning.
%training RLMs on a small set of the proposed structured reasoning significantly improves the performance on MT.

Looking into the reason behind the effectiveness of the structured reasoning, in \Cref{subsec:structured-reasoning-analysis}, and through controlled ablation studies with varying reasoning trace quality and target translation quality, we show that having higher quality reasoning traces (e.g., proposed structure reasoning) leads to performance gain on translation quality, even when paired with weaker final outputs. However, the quality of the final translation has a more positive role on the overall performance.

%we analyze the relative contributions of the proposed reasoning style and the final translation quality. In our setup, the target translation refers to the reference translation that the RLM is trained to generate, whose quality depends on the translation system that produces it.

We summarize our main contributions as follows:
\begin{itemize}
    \item We present a comprehensive evaluation and analysis of reasoning large language models for machine translation. We show that existing reasoning formats in RLMs, without revision, self-correction, or exploration, do not benefit MT. We also demonstrate that injecting reasoning traces with the same format, but higher-quality translations, does not constantly improve RLM's performance in MT (\Cref{sec:benchmarking}).
    \item We propose a structured reasoning framework tailored to machine translation, a designed synthetic reasoning generation pipeline that enforces iterative revision and self-correction. Training on the structured reasoning traces leads to a significant improvement in translation performance (\Cref{sec:structured-reasoning,sec:results}).
    \item We disentangle the roles of reasoning trace quality and final translation quality through controlled ablation studies on RLMs, showing that while high-quality structured reasoning traces provide performance improvements, the quality of the final translation is the more important factor in determining MT performance (\Cref{subsec:structured-reasoning-analysis})
\end{itemize}

% \tk{this section also sounds LLM generated. What do you mean for example by "effectiveness depends on its alignment with the underlying problem geometry". We should keep the core message, the introduction is already longer than usual}

% suggest that reasoning should not be treated as a task-agnostic capability that uniformly benefits all forms of generation, but as a task-structured process whose effectiveness depends on its alignment with the underlying problem geometry.
% This has broader implications for RLM design.
Together, these findings challenge the assumption that test-time scaling via longer generic reasoning traces will reliably improve performance across tasks, and instead point to the need for task-specific structural priors when generating synthetic reasoning data.
More generally, they motivate viewing reasoning not as a single transferable skill, but as a family of task-dependent deliberation strategies that must be learned and evaluated in a task-aware manner, with consequences for other generative tasks.

\begin{table*}[t]
\caption{Performance evaluation of reasoning language models on the WMT24++ benchmark using XCOMET-XL score. We observe that having reasoning traces hurts the translation quality across almost all languages and models.}
\vskip 0.1in
\centering
\small
\begin{tabular}{lccccccccc|c}
\toprule
\textbf{Model} & \textbf{en-ar} & \textbf{en-cs} & \textbf{en-fa} & \textbf{en-fr} & \textbf{en-hi} & \textbf{en-ja} & \textbf{en-ko} & \textbf{en-ru} & \textbf{en-zh} & \textbf{Avg.} \\
\toprule
\cmdareasoning{} (w)     & \worse{70.4}	& \worse{83.0}	& \worse{78.6} &                                  \worse{82.5}  & \worse{59.8}  & \worse{80.6} &                                  \worse{81.4}	& \worse{82.0}  & \worse{77.2} &                                  \worse{77.3} \\
\cmdareasoning{} (w/o)   & 72.6 & 83.5 & 81.2 & 83.0 & 60.4 & 80.7 & 81.7 &                              82.6 & 78.6 & \textbf{78.3} \\
\addlinespace
\claude{} (w)           & \worse{74.9} & \worse{84.9} & 81.5 & \worse{83.2}                             & \worse{64.4} & \worse{83.9} & \worse{83.5} & 
                              \worse{84.2} & \worse{80.7} & \worse{80.1} \\
\claude{} (w/o)         & 75.1 & 85.2 & \worse{81.0} & 83.4 & 64.6 & 84.0 &                             83.8 & 85.1 & 81.6 & \textbf{80.4} \\
\addlinespace
\deepseek{} (w)             & \worse{69.5} & \worse{82.5} & \worse{76.8} &                                    \worse{82.0} & \worse{61.0} & \worse{74.8} &                                    \worse{74.2} & \worse{83.0} & \worse{70.4} &                                    \worse{74.9} \\
\deepseek{} (w/o)           & 73.0 & 83.7 & 79.2 & 82.7 & 63.1 & 81.3 & 81.8 &                               83.8 & 79.3 & \textbf{78.7} \\
\addlinespace
\gemini{} (w)        & \worse{71.2} & \worse{83.9} & \worse{79.9} & 
                              \worse{82.5} & \worse{63.6} & \worse{81.9} & 
                              \worse{81.3} & \worse{83.4} & \worse{79.4} & 
                              \worse{78.6} \\
\gemini{} (w/o)      & 74.0 & 85.5 & 81.2 & 83.8 & 63.8 &                                              82.7 & 83.6 & 84.5 & 80.1 & \textbf{79.9} \\
\hline
\end{tabular}
\label{tab:benchmarking-reasoning-mt}
\end{table*}

\section{Related Work}
\paragraph{Reasoning Language Models}
Prompting language models to generate intermediate steps (commonly referred to as Chain-of-Thought (CoT)~\cite{wei-cot}) has yielded significant performance gains across complex, reasoning-demanding tasks like mathematics~\cite{cobbe2021gsm8k,yu-etal-2025-chain}, logic reasoning~\cite{sprague2025to}, and programming~\cite{min2024beyond,wang-etal-2025-exploracoder}. 
In a similar direction, it has been shown that reasoning capabilities can also emerge through pure reinforcement learning (RL) training~\cite{Guo_2025,mistralai2025magistral}.
Beyond performance gains, recent studies have also looked into reasoning trajectories generated by RLMs, investigating their structures and features and validating the intermediate steps~\cite{marjanovic2026deepseekr,liao2025lostbeginningreasoning,liu-etal-2024-lost,zhang-etal-2024-evaluating}.

% Removing the below, as we don't distill.
%% \paragraph{Knowledge distillation in RLMs}
%% Distilling reasoning knowledge from a strong teacher model into smaller or non-reasoning LLMs has become a primary setup to enhance the reasoning capabilities of LLMs~\cite{shridhar-etal-2023-distilling,meta_llama4_2025,gu2025miniplm}.
%% Prior work has shown that knowledge distillation is more effective than an RL setup in improving performance~\cite{Guo_2025}.
%% The success has led to the creation of many distilled reasoning datasets, such as S1k~\cite{muennighoff-etal-2025-s1}, LIMO~\cite{ye2025limoreasoning}, and OpenThoughts~\cite{guha2025openthoughtsdatarecipesreasoning}, a few among others.
%% However, the trade-off between the quality and quantity of distilled reasoning traces and their correlation with performance gains remains an open question, with contradictory observations~\cite{li2025naturalthoughtsselectingdistillingreasoning}. 
\paragraph{Reasoning for MT}
Several studies have explored CoT prompting for machine translation, reporting remarkable gains on the performance~\cite{nguyen-xu-2025-reasoning,wang-etal-2025-drt}. Moreover, it has been  that multi-step translation focusing on different aspect of translations in each step (e.g., drafting or fluency)~\cite{briakou-etal-2024-translating}, as well as simple iterative translation refinements~\cite{wu-etal-2025-please}, boost the translation quality.
More recently, a similar work to ours has evaluated RLMs on MT showing that explicit thinking tokens do not improve translation quality, a finding that aligns with our observations~\cite{zebaze2025llmreasoningmachinetranslation}. Furthermore, they have also reported that fine-tuning with synthetic CoT explanations does not outperform standard input-output fine-tuning on MT. This conclusion that is not supported by our experimental results.

\par
In a nutshell, our work differs from prior studies in a few key aspects: (1) we conduct a systematic evaluation and in-depth analyses on the role of reasoning on machine translation performance in RLMs showing the ineffectiveness of thinking tokens on MT and investigating the reasons behind it. (2) We design a dynamic reasoning generation pipeline to synthetically generate helpful reasoning traces for MT, and we show a solid performance improvement using our structured reasoning framework. (3) We carry out an ablation study to investigate the contribution of reasoning traces quality compared to the final translation quality in the synthetic dataset, showing that both positively affect the performance while the final translation quality have a stronger role on RLMs performance.

\section{Reasoning Capability Inspection}
\label{sec:benchmarking}
Reasoning skills have gained special attention in recently developed large language models. It has been shown that explicit reasoning, also known as \textit{thinking tokens}, through decomposition, self-verification, and iterative refinement substantially improves LLMs' performance across complex tasks, such as mathematical problems or code generation. However, the extent to which the reasoning capabilities benefit open-ended generation tasks like machine translation (MT) remains less-explored. 

To investigate the role of reasoning on MT, we begin with evaluating four open and closed-weights reasoning language models, \cmdareasoning{}~\cite{cohere2025commandaenterprisereadylarge}, \claude{}~\cite{Anthropic_Claude_4_System_Card_2025}, \deepseek{}~\cite{Guo_2025}, and \gemini{}~\cite{comanici2025gemini25pushingfrontier}. We evaluate these models on nine language pairs from the WMT24++ benchmark and compare the quality of generated translation with and without reasoning using the XCOMET-XL metric~\citep{guerreiro-etal-2024-xcomet}, a state-of-the-art metric for automatic evaluation \citep{freitag-etal-2024-llms}. Additional experimental details are provided in \Cref{app:experimental-setups}.
%\sv{Reviewer question: Why did you select these specific models?}
%\sr{I think whatever we pick as model, a reviewer might ask for another one.}

\Cref{tab:benchmarking-reasoning-mt} summarizes the results. 
At a high-level glance and looking into the average performance, we observe that all the reasoning models consistently have a significantly higher performance without reasoning. 
The pattern holds for all the individual language pairs as well, except for Farsi in \claude{} where we see that reasoning improves the translation quality. 
Our findings show that while it has been widely reported in previous studies that reasoning abilities help LLMs to improve their performance on a variety of tasks~\cite{pham2025sealqaraisingbarreasoning,huang2025englishcentrictrainingreinforcementlearning,son2026pushingmultilingualreasoningmodels}, it has an opposite effect on the translation performance.  
 % This aligns with prior findings that reasoning is not universally helpful in all scenarios. 

To further investigate the characteristics of reasoning traces in the machine translation task, we conduct a series of qualitative and quantitative analyses, focusing on the open-weights models, in the following sections.

\subsection{Structural Analysis}
\label{subsec:structural_analysis}
\begin{table}[t]
\centering
\small
\caption{Average number of reasoning paths in a generated reasoning trace per model, aggregated over the selected WMT24++ language pairs. All the RLMs show a high degree of linearity in their reasoning process.}
\label{tab:num-reasoning-path}
\vskip 0.1in
\begin{tabular}{lcc}
\toprule
\textbf{Model} & \textbf{Avg. \# Reasoning Paths} & \textbf{Std.} \\
\midrule
\cmdareasoning{}      & \textbf{1.56} & 0.23 \\
\claude{}            & 0.006 & 0.0 \\
\deepseek{}       & 0.107 & 0.16 \\
\gemini{}         & 0.012 & 0.0 \\

\bottomrule
\end{tabular}

\end{table}

To better understand why having reasoning traces fails to improve MT performance, we analyze the structure of reasoning traces for the translation task.
\Cref{app:reasoning-example} shows an example of a generated reasoning trajectory by \deepseek{}.

Looking into the generated reasoning traces, we notice that, unlike other reasoning-demanding tasks (like math problems), where the RLM explores multiple scenarios, evaluates intermediate steps, and can backtrack~\cite{an2025dont}, MT reasoning style exhibits a linear structure. The RLMs split the source text into segments, translate each segment in isolation without considering possible alternative translations, and proceed sequentially without revisiting earlier translations. 

To put this in a numerical perspective, we calculate the number of distinct reasoning paths (i.e., exploring different possible translations) that an RLM explores in its generated reasoning traces for MT. Following previous studies~\cite{an2025dont,cuesta-ramirez-etal-2025-lrms}, we use linguistic cues,  \textit{Wait} and \textit{Alternatively} as a signal for a reasoning path and count their occurrences.

We report the average number of different reasoning paths within the whole reasoning trace in \Cref{tab:num-reasoning-path}. The numerical results confirm our qualitative observation that MT reasoning traces are highly linear, lacking exploration or self-correction in the reasoning process. Among the selected models, \cmdareasoning{} has the highest number of reasoning paths in the reasoning traces on average, meaning that the model explores alternative translation reformulation for $1.56$ number of segments in the given text on average. It is worth mentioning that a given text could have multiple segments. 
% evaluate how comprehensively RLMs reason for machine translation task. To this end, we calculate the percentage of the given source text is covered in the reasoning trace. 
% Two important reasoning characteristics are notably missing. First, reasoning trajectories do not include steps that integrate segment-level translations into a unified translation. Second, the LRM rarely revises or verifies earlier translations. 

\subsection{Reasoning Injection}
\label{subsec:reasoning_injection}

\begin{figure}
    \centering
    \includegraphics[width=0.99\linewidth]{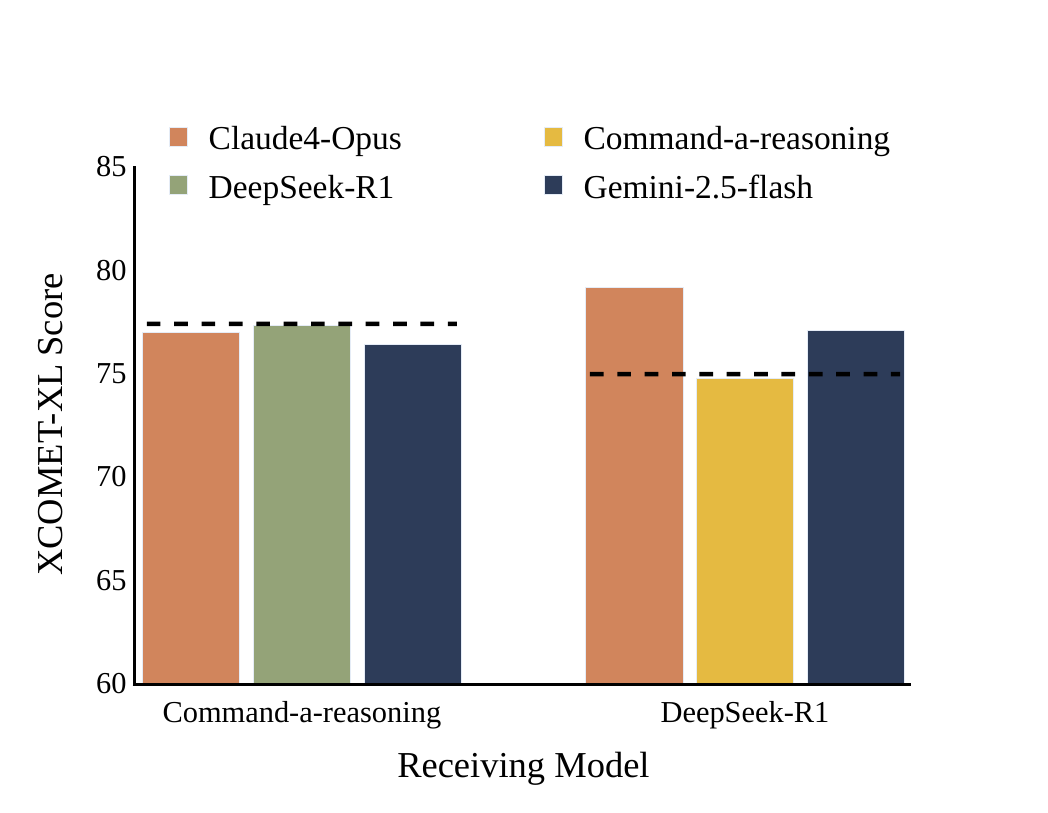}
    \caption{The performance of open-weights RLMs when their reasoning traces are generated and injected from other models. The dash-line represents the baseline performance of the receiving models, and the injecting models are presented at the top. As can be seen, the strength of the injecting model used to generate reasoning traces does not correlate with final translation quality across different injecting and receiving models.}
    \label{fig:reasoning-injection}
\end{figure}
In our primary experiment, we have shown that having reasoning does not bring performance gain on the machine translation task for RLMs, and this pattern is consistent across all the selected models. Nevertheless, the absolute translation performance varies across models, with \claude{} achieving the highest overall performance. In this part, we investigate whether a higher-quality reasoning trace generated by a stronger model on MT can provide other models with a hint to improve their performance. 

We conduct a controlled reasoning injection experiment. For each translation example, we generate a reasoning trace using an injecting model (e.g., \claude{}), insert it into another receiving model (e.g., \cmdareasoning{}), and let the receiving model generate the final translation using the provided reasoning trace. To have full control over the reasoning traces between \textit{thinking tokens}, we limit ourselves to the open-weights models, namely \cmdareasoning{} and \deepseek{}, as the receiving models, while using all four models as injecting ones to generate reasoning traces.

The results are presented in \Cref{fig:reasoning-injection}. First, we don't observe a notable performance drop in the receiving models, indicating that RLMs are not biased toward their own reasoning style for machine translation. Secondly, although reasoning traces from stronger injecting models, such as \claude{}, lead to performance improvement of \deepseek{}, this pattern does not hold for \cmdareasoning{}. In other words, reasoning traces generated by stronger models do not consistently yield higher translation quality across different receiving models.

In summary, we find that the reasoning styles for machine translation differ from other reasoning-demanding tasks. In their current format, translation reasoning trajectories follow a linear process and rarely include translation revision or exploration of possible translation options, and even higher-quality reasoning traces from stronger models cannot benefit other RLMs to enhance their translation performance, suggesting the ineffectiveness of the existing reasoning format in the MT task.

% Complementary to our observations, there are findings that models are able to reason when prompted as a chain-of-thoughts. ~\citet{briakou-etal-2024-translating} have shown that a step-by-step approach including multiple translation versions significantly boosts translation quality. Similarly,~\citet{wu-etal-2025-please} demonstrate that iterative revisions using LLMs lead to notable gains in translation performance, and \citet{kocmi-etal-2025-command} employ multi-turn, step-by-step translation to boost translation system performance.
Building on our findings and inspired by previous studies on the effectiveness of step-wise translation methods, in the next section, we propose a framework for creating dynamic, step-by-step reasoning traces for machine translation.

\section{Structured Reasoning Trace}
\label{sec:structured-reasoning}
We design a synthetic data generation pipeline to train reasoning skills beneficial for machine translation in reasoning language models. Building on our findings in the previous section showing that the state-of-the-art reasoning language models do not generate useful reasoning traces for machine translation and the success of chain-of-thoughts techniques used in MT~\cite{nguyen-xu-2025-reasoning,10.1162/TACL.a.25,briakou-etal-2024-translating}, we aim to create structured reasoning traces, teaching the model how to do step-by-step reasoning, and improve translation through successive refinements. 
%Therefore, each training example consists of a source document, structured reasoning trajectory, and a final translation in the target language. 
In the following subsections, we describe our data generation process, quality check of reasoning traces, and the construction of reasoning templates.

\subsection{Multi-turn Translation Annotation}
\begin{figure*}[tb]
\centering
\begin{tcolorbox}[
  width=\textwidth,
  enhanced,
  colback=white,
  colframe=FrameGray,
  boxrule=0.5pt,
  arc=1mm,
  left=6pt,right=6pt,top=5pt,bottom=6pt,
  fontupper=\ttfamily\small,
  title=\textbf{Dynamic Structured Reasoning Template},
  coltitle=HeaderFG,
  colbacktitle=HeaderBG,
  colframe=FrameGray,
  boxed title style={
    sharp corners,
    boxrule=0pt,
    left=6pt,right=6pt,top=3pt,bottom=3pt
  },
]
\textbf{Step 1 --- Initial Draft}\\
Let me start by producing an initial, faithful translation of the source text into \{target\_language\}, which will serve as the foundation for later refinements.\\
\{initial\_draft\}\\

\vspace{3pt}\hrule\vspace{3pt}

\textbf{Step 2 --- Adequacy}\\
In this step, I focus on adequacy, ensuring that the full meaning of the challenging sentences is conveyed accurately and completely in \{target\_language\}.\\
\{challenging\_sentences\_adequacy\}\\

\vspace{3pt}\hrule\vspace{3pt}

\textbf{Step 3 --- Fluency}\\
At this stage, I refine the challenging sentences, ensuring they read smoothly and naturally in \{target\_language\}, as if originally written in it.\\
\{challenging\_sentences\_fluency\}\\

\vspace{3pt}\hrule\vspace{3pt}

\textbf{Step 4 --- Final Translation}\\
Finally, I bring together all the work from the previous steps, addressing any remaining issues or translation errors to produce a polished final version in \{target\_language\}.\\

\end{tcolorbox}

\caption{An example of the dynamic template used to generate structured reasoning traces. Curly-braced fields denote placeholders. Steps 2 and 3 are applied only to selected challenging sentences, those sentences that benefited the most from steps 2 and 3 with the defined MetricX margin; all other segments are discarded from the initial draft.}
\label{fig:dynamic-reasoning-template}
\end{figure*}

We begin with a high-quality monolingual corpus of documents spanning 23 source languages, prepared to be translated. All the documents pass quality control and difficulty sampling~\cite{proietti-etal-2025-estimating,kocmi-etal-2025-command}. Following~\citet{briakou-etal-2024-translating}, for each source document, we generate a sequence of four translation refinements using an offline, step-wise procedure employing the \cmdatranslate{}~\cite{kocmi-etal-2025-command}. Each step focuses on a distinct and important aspect of translation.

%\tk{a point we should mention is that we are discarding samples that can't keep up the line alignment, which is about 10\% of all samples, therefore the 320k is much higher number. Done}

\begin{enumerate}
    \item \textbf{Initial Draft:} \cmdatranslate{} is prompted to generate an initial draft intended to be rough and as a foundation for the next steps of translation.
    \item \textbf{Adequacy Enhancement:} Given the draft translation, the model is asked to revise the translation with an explicit emphasis on semantic accuracy.
    \item \textbf{Fluency Refinement:} In this step, the LLM improves the readability and fluency of the adequate translation in the target language. 
    \item \textbf{Final Translation:} In this final stage, \cmdatranslate{} considers all the previous translations, resolving remaining errors, and generates its best possible translation. 
\end{enumerate}
We consider this four-step trajectory as a structured approximation of a useful \textit{reasoning process} for translation. In this step, we discard samples whose translated documents are not line-aligned with the source documents, a requirement for the follow up quality estimation, resulting in $320k$ high-quality samples. Detailed prompts used for each step of translations are presented in \Cref{app:translation-prompts}.

\subsection{Quality-based Data Selection}
Having the multi-turn translations for all $320k$ examples, we notice that only a small subset shows meaningful improvements across the refinement steps. We quantify the extent of improvement for each intermediate translation using MetricX~\cite{juraska-etal-2024-metricx}.
%To quantify this, we evaluate each intermediate translation using metricX. 
Since the MetricX has limited context window, we evaluate all lines in isolation, averaging them into the document-level score.

We observe that less than $10\%$ of examples show substantial quality gains (e.g., more than $0.5$ MetricX improvement, technically meaning half a minor error in the MQM terminology~\cite{burchardt-2013-multidimensional}) across different steps. This is expected as SOTA MT systems like \cmdatranslate{} have remarkably high performance and often translate well in a single pass, meaning most of the translation examples are not challenging for them. Therefore, we calculate the MetricX for each segment of a given document and consider the average MetricX score as the whole document score. As stated before, we keep only examples where the improvement between the initial draft and the final translation is at least $0.5$ MetricX score. We find the filtering step critical for improving RLMs performance. 
%\tk{highlight that we score them segment by segment and then use the aggregated document score}
This filtering yields a collection of 28k high-quality samples, each demonstrating a non-trivial translation. A breakdown of language-pair distribution for this subset is provided in \Cref{fig:lang-dist}. 

\subsection{Constructing Reasoning Traces}
Having instances where step-by-step translation is beneficial, we next convert the multi-step outputs into reasoning traces. We hypothesize that training the model with a step-by-step style of reasoning helps it use this structure for reasoning in the machine translation task. We later verify our hypothesis through empirical experiments.
%that we later use to teach a helpful reasoning process to the model for the machine translation task.

\paragraph{Static Template}
The first way to create the reasoning trajectories is to chain the four translations into a single trace. We treat a set of human-generated sentences as connectives to connect four translations. To diversify the reasoning structure, for each step translation, we randomly insert one of the predefined sentences into the reasoning trace reflecting a \textit{natural thinking flow} in the model.
%Between consecutive steps, we randomly insert one of the ten predefined connective sentences to reflect a \textit{natural thinking process} of the model. 
An illustration of this template is provided in \Cref{app:static-reasoning-template}. A concern raised by this template is that the trace length scales with document size, which may introduce inefficiencies for long inputs.

\paragraph{Dynamic Template}
As mentioned above, having all four steps for the whole document is not efficient, as not all the segments of a document need translation through iterative reasoning. To enhance the efficiency of the reasoning traces, we develop a dynamic template that, after the initial draft, includes reasoning only for \textit{challenging} segments. We consider sentences challenging that the translation quality between the initial drafting step and the adequacy or fluency stage improves by at least 1 MetricX point. 
More specifically, in this template, we include the initial draft translation of the entire document. Then, we drop the translation of segments without meaningful improvement in the fluency and adequacy steps from the reasoning trace. We also discard the final translation from the trace to reduce repetition. \Cref{fig:dynamic-reasoning-template} shows an example of the resulting dynamic template strategy. More details on dynamic templates are presented in \Cref{app:structured-reasoning-template}.
% \sv{You should report the accuracy/XCOMET-XL change, but also *very important* the trigger rate for reasoning traces (in how many cases did they actually end up being "simple" and in how many "complex"? Do you have a held out test set with gold-standard annotations to show how well the model learns to dynamically adapt the traces?}
% \sr{In line 177, we explain only 10\% of the examples are complex. about the last question, you mean what is the percentage of the selected sentences for step 2 and 3?}
\subsection{Final Training Examples}
Each training sample in our dataset consists of the triple \textit{(source text, reasoning trace, target text)} in which reasoning traces are constructed using the explained dynamic template, and the target text is adopted as the \cmdatranslate{} final-step translation. 
The resulting 28k data provides explicit, structured reasoning for translation, which we use to train reasoning models to carry out step-by-step reasoning and improve their reasoning capability for machine translation.

\begin{table*}[htbp]
\centering
\small
\caption{The systematic performance evaluation of different reasoning-based translation systems. At the top, we have the \cmdareasoning{} as the backbone model and Direct Translations, where the base model is trained on MT without reasoning. The structured reasoning is the base model trained on our proposed structured reasoning. Injected Reasoning represents a comparison between the performance of the base model trained on injected reasoning from different reasoning models. At the bottom, a comparison between different training sizes of structured reasoning}
\vskip 0.1in
\begin{tabular}{lcccccccccc}
\hline
\textbf{Setup} & \textbf{en-ar} & \textbf{en-cs} & \textbf{en-fa} & \textbf{en-fr} & \textbf{en-hi} & \textbf{en-ja} & \textbf{en-ko} & \textbf{en-ru} & \textbf{en-zh} & \textbf{Avg.} \\
\hline
\textit{Baselines}  &  &  &  &  &  &  &  &  &  &  \\

\hspace{1em}Base model             & 70.4  & 83.0  & 78.6 & 82.5 & 59.8 
                       & 80.6	& 81.4	& 82.0 & 77.2 & 77.3\\
                       
\hspace{1em}Direct Translation     & 76.7 & 86.3 & 82.7 & 85.7 & 66.2 
                       & 83.7 & 84.5 & 85.5 & 81.9 & 81.5 \\
\midrule
Structured Reasoning $_{\textit{ours}}$   & 77.8  & 87.1  & 83.6 & 86.4 & 66.3 
                       & 83.8  & 84.7  & 86.2 & 82.1 & \textbf{82.0} \\

\midrule
\textit{Injected Reasoning} &  &  &  &  &  &  &  &  &  &  \\
\hspace{1em}\cmdareasoning{}           & 76.9 & 86.7 & 82.9 & 85.7 & 66.1 & 83.4 & 84.5 & 85.8 & 82.3 & 81.6 \\
\hspace{1em}\deepseek{}                 & 76.5 & 86.5 & 83.1 & 85.7 & 66.1 & 83.7 & 84.5 & 85.6 & 82.3 & 81.6 \\

\midrule
\textit{Structured Reasoning} &  &  &  &  &  &  &  &  &  &  \\
\hspace{1em}1k Samples         & 74.3 & 80.2 & 81.0 & 83.5 & 60.9 & 80.0 & 80.6 & 83.8 & 79.4 & 78.2 \\
\hspace{1em}10k Samples                & 77.6 & 86.7 & 83.4 & 85.9 & 65.9 & 83.6 & 84.7 & 86.1 & 81.8 & 81.7 \\

% Radom Sampling - 320k   & -  & -  & - & - & - 
%                        & - & -  & - & - & - \\
\hline
\end{tabular}

\label{tab:main-results}
\end{table*}

\section{Experimental Setup}
As a post-training stage, we do supervised fine-tuning (SFT) on the \cmdareasoning{} model with 111B parameters using our curated 28k translation-reasoning examples. We pick this model as it is among the strongest open-weights reasoning models. All the experiments are conducted with a maximum sequence length of $32768$ tokens, and we evaluate the impact of different data selection strategies by training models on subsets of $1k$, $10k$, and the full $28k$ dataset. 
% To understand the role of data quality, we additionally compare against models trained on randomly selected samples from the original $320k$ unfiltered corpus.
% \sv{why \cmdareasoning{} and not others? You should explain that this is Cohere's flagship/best reasoning model at the time of the experiments --> paper is Cohere-affiliated --> default choice}

Unless otherwise noted, models trained on $28k$ samples are finetuned for $4$ epochs. Models trained on $1k$ and $10k$ subsets are trained for $8$ and $6$ epochs, respectively. We use a peak learning rate of $1\times 10^{-5}$, decayed to $1\times 10^{-6}$, with additive weight decay of $0.1$.  All the SFT runs are done on $64$ first class GPUs. 

% \tk{all these parameters could be on one spot, maybe together with the evaluation parameters, could even go to the appendix as it is less important for the flow of the paper}

\subsection{Baselines}
To isolate the contribution of the designed structured reasoning traces, we compare our approach to the following baselines:
\paragraph{Direct Translation}
To disentangle the effect of additional MT supervision from explicit reasoning, we finetune the same base model using only direct translations without reasoning traces. This setup is the strong baseline as it shows the impact of training off high quality MT system. We employ a set of ten templates within the thinking tokens, instructing the model to skip step-by-step reasoning. The used templates are available in \Cref{app:direct-translation-template}.

\paragraph{Injected Reasoning Traces} As a contrastive ablation, we train the same base model on reasoning traces generated for the same 28k examples using \cmdareasoning{} and \deepseek{}. Note that only the reasoning traces are generated from these models, and the \textit{target text} translation is kept from the \cmdatranslate{} as in the other experiments. This allows us to assess the role of structured reasoning traces compared to linear reasoning trajectories. 

To ensure a fair comparison, we use the same hyperparameters as the structured reasoning for the baselines.

\section{Results}
\label{sec:results}

We evaluate all models on the WMT24++ benchmark~\cite{deutsch-etal-2025-wmt24}, focusing on nine diverse language pairs. Translation quality is evaluated using the XCOMET-XL metric~\cite{guerreiro-etal-2024-xcomet}. For all evaluations, we set the decoding temperature to $0$ to minimize variance and maximize reproducibility. 

\Cref{tab:main-results} summarizes the overall results. In the first block of the table, we compare the performance of our structured reasoning approach against both the base model and the direct translation finetuning setup. Direct Translation is a strong baseline as we ensure the performance improvement is not solely due to additional high-quality training data. 
Across all language pairs, structured reasoning shows substantial performance improvements over these baselines, indicating that the structured dynamic template is effective in developing a reasoning process that meaningfully supports translation quality. It is worth mentioning that $0.5$ XCOMET-XL score improvement is considered as noticeable by human as +1 BLEU improvement \citep{kocmi-etal-2024-navigating}.

The middle part of the table compares our synthetic reasoning traces with the models trained on the trajectories generated by \deepseek{} and \cmdareasoning{}. Comparing these models against the direct translation baseline, we see that the injected reasoning does not offer additional benefits. This finding is consistent with our earlier analysis in Section~\ref{sec:benchmarking}, where we show that the existing reasoning capabilities of reasoning models are beneficial for the machine translation task.

Finally, in the bottom part of \Cref{tab:main-results}, we conduct an ablation on the size of the training data to investigate the relation between the training size and performance improvement. We see a monotonically increasing trend as we use more data for training, aligning with~\citet{li2025naturalthoughtsselectingdistillingreasoning} observations on positive effect of training size.  Interestingly, about one-third of the training data surpasses the direct translation baseline, highlighting the . 
% Table ~\ref{tab:main-results} presents an overview of the results. In the first part of the table, we compare the performance of the structured reasoning to the base model and Direct translation setup. Comparing to both baselines, structured reasoning achieves significantly higher performance across all language pairs, suggesting the effectiveness of designed dynamic template. 

% We also compare the quality of our synthetic reasoning traces with reasoning trajectories distilled from \deepseek{} and \cmdareasoning{} in the middle part of the table. Comparing the performance of this category with the direct translation experiment, we clearly see there is not benefit out the reasoning of these models. This is aligns with our previous finding in Section~\ref{sec:benchmarking}

% In the third part the Table, we investigate the role of training data size using the structured reasoning traces on the performance. Following the scaling low, we observe consistent improvement in the performance as we increase the size of training data. 

\begin{table*}[htbp]
\centering
\small
\caption{Comparing the role of reasoning quality and final translation quality on the performance. \textit{R-Src} represents the injecting model of reasoning traces. Each row in a block represents the model that generated the final translations. As shown, both the quality of reasoning traces and final translations contribute positively to performance across languages, indicating that reasoning traces from stronger models improve performance and that higher-quality final translations in the training data lead to better performance.}
\vskip 0.1in
\begin{tabular}{lcccccccccc}
\hline
\textbf{Setup} & \textbf{en-ar} & \textbf{en-cs} & \textbf{en-fa} & \textbf{en-fr} & \textbf{en-hi} & \textbf{en-ja} & \textbf{en-ko} & \textbf{en-ru} & \textbf{en-zh} & \textbf{Avg.} \\
\hline
\textit{R-Src: \cmdareasoning{}}  &  &  &  &  &  &  &  &  &  &  \\

\hspace{1em} \cmdareasoning{}            & 73.4  & 83.4  & 79.5 & 83.4 & 60.3 
                       & 80.4	& 81.0	& 82.6 & 78.6 & 78.1\\
                       
\hspace{1em} \cmdatranslate{}     & 76.9 & 86.7 & 82.9 & 85.7 & 66.1 & 83.4 & 84.5 & 85.8 & 82.3 & 81.6 \\

\midrule
\textit{R-Src: \deepseek{}} &  &  &  &  &  &  &  &  &  &  \\
\hspace{1em} \deepseek{}           & 72.0 & 82.3 & 77.9 & 81.9 & 59.9 & 80.2 & 81.0 & 81.9 & 76.9 & 77.1 \\
\hspace{1em} \cmdatranslate{}               & 76.5 & 86.5 & 83.1 & 85.7 & 66.1 & 83.7 & 84.5 & 85.6 & 82.3 & 81.6 \\

\midrule
\textit{R-Src: Structured reasoning} &  &  &  &  &  &  &  &  &  &  \\
\hspace{1em} \deepseek{}           & 74.3 & 84.0 & 80.4 & 83.1 & 62.2 & 82.1 & 82.6 & 83.3 & 76.4 & 78.7 \\
\hspace{1em} \cmdareasoning{}               & 74.4 & 84.3 & 81.0 & 83.4 & 61.6 & 81.9 & 82.4 & 83.3 & 80.0 & 79.1 \\

\hline
\end{tabular}
\label{tab:reasoning-quality}
\end{table*}
\subsection{Reasoning vs. Translation Quality}
\label{subsec:structured-reasoning-analysis}

To further examine the relative contribution of reasoning trace quality and target translation quality, we design a controlled ablation study using our $28k$ samples. 
As we mentioned earlier, each training example includes a triple of \textit{(source text, reasoning trace, target text)}. Therefore, the quality of both the reasoning trace and the target text affects the model performance. 

In this analysis, we explicitly vary these two factors independently. We consider three sources of reasoning traces: reasoning trajectories generated from \cmdareasoning{}, \deepseek{}, and our proposed structured reasoning. Separately, we consider three sources of target translations: translations generated by \cmdareasoning{}, \deepseek{}, and the stronger model \cmdatranslate{}. By pairing each reasoning trace source with each target translation source, we construct multiple training datasets in which the reasoning traces and target translations may originate from different models. We then run SFT on the same base model using each dataset and evaluate the resulting model's translation performance.

The results are reported in \Cref{tab:reasoning-quality}. Each block is dedicated to a reasoning trace source, and each row within a block presents the target translation source model.
Focusing on the top two blocks, we observe that for a fixed reasoning source, using target translations generated by the stronger model (i.e., \cmdatranslate{}) consistently leads to substantially higher performance than using target translations from the other models. Interestingly, for the \cmdatranslate{}'s target translation set, we do not observe meaningful performance differences between reasoning traces generated by \cmdareasoning{} and those by \deepseek{}.

In the bottom block of \Cref{tab:reasoning-quality}, we evaluate structured reasoning paired with target translations from \cmdareasoning{}. This configuration yields higher performance than using \deepseek{} translations. 
This is expected, as in \Cref{tab:benchmarking-reasoning-mt}, we have shown that \cmdareasoning{} is a better model than \deepseek{} on the translation task.
However, the gains remain notably smaller than those obtained when using \cmdatranslate{} as the target translation (\Cref{tab:main-results}). 

Overall, these results indicate that target translation quality has a substantially larger impact on downstream performance than the quality of the associated reasoning traces in machine translation. This suggests that improvements in reasoning traces alone are insufficient to compensate for weaker target translations.

\section{Conclusions}
% \tk{We could add the findings how the structured reasoning was spilling into non MT tasks, that was really interesting findings. But may be just a sidenote}
In this paper, we conduct an in-depth study on the role of reasoning capabilities of reasoning language models in machine translation. Despite the substantial benefits that reasoning has provided for most reasoning-intensive tasks, our experiments show that current reasoning degrades translation quality across RLMs and languages. Through detailed analyses, we notice that the structure of current reasoning traces is highly linear, lacking key features such as revision, self-correction, and exploration of possible alternatives. 
Building on our findings, we propose a structured reasoning framework specifically designed for MT that dynamically adapts the length of reasoning to the difficulty of translation. We show that post-training an RLM on a small set of structured reasoning substantially improves the translation performance.
Finally, we investigate the underlying factors behind the effectiveness of structured reasoning by disentangling the contributions of reasoning trace quality and final translation quality. Our results indicate that while both factors affect MT performance, the quality of final translations plays a dominant role.
\section*{Impact Statement}

In this work, we present in-depth analyses on the reasoning capabilities of RLMs for MT. We propose a structured reasoning generation pipeline to enhance translation performance in RLMs. We do not foresee any significant negative societal impacts directly resulting from this work.

% \tk{Can you check that limitation section is not mandatory? It usually is and it is a good science to have it. We should definitely list the limitations of automatic evaluation, the lack of human evaluation.}
\section*{Limitations}
In this work, we use WMT24++ as our primary evaluation benchmark, focusing on 9 language pairs. While we have tried to select a diverse set of language pairs, adding more could strengthen our conclusions and findings. Moreover, we solely employ automatic metrics for evaluation and to investigate the role of reasoning on the translation quality. While costly, incorporating human evaluation could provide additional support for our findings.

\newpage
\bibliography{anthology-1,anthology-2,custom}

\begin{thebibliography}{44}
\providecommand{\natexlab}[1]{#1}
\providecommand{\url}[1]{\texttt{#1}}
\expandafter\ifx\csname urlstyle\endcsname\relax
  \providecommand{\doi}[1]{doi: #1}\else
  \providecommand{\doi}{doi: \begingroup \urlstyle{rm}\Url}\fi

\bibitem[An et~al.(2025)An, Wang, Zhou, and Hsieh]{an2025dont}
An, S., Wang, R., Zhou, T., and Hsieh, C.-J.
\newblock Don{\textquoteright}t think longer, think wisely: Optimizing thinking dynamics for large reasoning models.
\newblock In \emph{The Thirty-ninth Annual Conference on Neural Information Processing Systems}, 2025.
\newblock URL \url{https://openreview.net/forum?id=nxnBaaRLnz}.

\bibitem[Anthropic(2025)]{Anthropic_Claude_4_System_Card_2025}
Anthropic.
\newblock {System Card: Claude Opus 4 \& Claude Sonnet 4}.
\newblock Technical report, Anthropic, May 2025.
\newblock URL \url{https://www.anthropic.com/claude-4-system-card}.

\bibitem[Briakou et~al.(2024)Briakou, Luo, Cherry, and Freitag]{briakou-etal-2024-translating}
Briakou, E., Luo, J., Cherry, C., and Freitag, M.
\newblock Translating step-by-step: Decomposing the translation process for improved translation quality of long-form texts.
\newblock In Haddow, B., Kocmi, T., Koehn, P., and Monz, C. (eds.), \emph{Proceedings of the Ninth Conference on Machine Translation}, pp.\  1301--1317, Miami, Florida, USA, November 2024. Association for Computational Linguistics.
\newblock \doi{10.18653/v1/2024.wmt-1.123}.
\newblock URL \url{https://aclanthology.org/2024.wmt-1.123/}.

\bibitem[Cobbe et~al.(2021)Cobbe, Kosaraju, Bavarian, Chen, Jun, Kaiser, Plappert, Tworek, Hilton, Nakano, Hesse, and Schulman]{cobbe2021gsm8k}
Cobbe, K., Kosaraju, V., Bavarian, M., Chen, M., Jun, H., Kaiser, L., Plappert, M., Tworek, J., Hilton, J., Nakano, R., Hesse, C., and Schulman, J.
\newblock Training verifiers to solve math word problems.
\newblock \emph{arXiv preprint arXiv:2110.14168}, 2021.

\bibitem[{Cohere Team}(2025)]{cohere2025commandaenterprisereadylarge}
{Cohere Team}.
\newblock Command a: An enterprise-ready large language model, 2025.
\newblock URL \url{https://arxiv.org/abs/2504.00698}.

\bibitem[Cuesta-Ramirez et~al.(2025)Cuesta-Ramirez, Beaussant, and Mounsif]{cuesta-ramirez-etal-2025-lrms}
Cuesta-Ramirez, J., Beaussant, S., and Mounsif, M.
\newblock {LRM}s are not thinking straight: Unreliability of thinking trajectories.
\newblock In Wartena, C. and Heid, U. (eds.), \emph{Proceedings of the 21st Conference on Natural Language Processing (KONVENS 2025): Long and Short Papers}, pp.\  27--32, Hannover, Germany, September 2025. HsH Applied Academics.
\newblock URL \url{https://aclanthology.org/2025.konvens-1.3/}.

\bibitem[{DeepSeek Team}(2025)]{Guo_2025}
{DeepSeek Team}.
\newblock Deepseek-r1 incentivizes reasoning in llms through reinforcement learning.
\newblock \emph{Nature}, 645\penalty0 (8081):\penalty0 633–638, September 2025.
\newblock ISSN 1476-4687.
\newblock \doi{10.1038/s41586-025-09422-z}.
\newblock URL \url{http://dx.doi.org/10.1038/s41586-025-09422-z}.

\bibitem[Deutsch et~al.(2025)Deutsch, Briakou, Caswell, Finkelstein, Galor, Juraska, Kovacs, Lui, Rei, Riesa, Rijhwani, Riley, Salesky, Trabelsi, Winkler, Zhang, and Freitag]{deutsch-etal-2025-wmt24}
Deutsch, D., Briakou, E., Caswell, I.~R., Finkelstein, M., Galor, R., Juraska, J., Kovacs, G., Lui, A., Rei, R., Riesa, J., Rijhwani, S., Riley, P., Salesky, E., Trabelsi, F., Winkler, S., Zhang, B., and Freitag, M.
\newblock {WMT}24++: Expanding the language coverage of {WMT}24 to 55 languages {\&} dialects.
\newblock In Che, W., Nabende, J., Shutova, E., and Pilehvar, M.~T. (eds.), \emph{Findings of the Association for Computational Linguistics: ACL 2025}, pp.\  12257--12284, Vienna, Austria, July 2025. Association for Computational Linguistics.
\newblock ISBN 979-8-89176-256-5.
\newblock \doi{10.18653/v1/2025.findings-acl.634}.
\newblock URL \url{https://aclanthology.org/2025.findings-acl.634/}.

\bibitem[Freitag et~al.(2024)Freitag, Mathur, Deutsch, Lo, Avramidis, Rei, Thompson, Blain, Kocmi, Wang, Adelani, Buchicchio, Zerva, and Lavie]{freitag-etal-2024-llms}
Freitag, M., Mathur, N., Deutsch, D., Lo, C.-K., Avramidis, E., Rei, R., Thompson, B., Blain, F., Kocmi, T., Wang, J., Adelani, D.~I., Buchicchio, M., Zerva, C., and Lavie, A.
\newblock Are {LLM}s breaking {MT} metrics? results of the {WMT}24 metrics shared task.
\newblock In Haddow, B., Kocmi, T., Koehn, P., and Monz, C. (eds.), \emph{Proceedings of the Ninth Conference on Machine Translation}, pp.\  47--81, Miami, Florida, USA, November 2024. Association for Computational Linguistics.
\newblock \doi{10.18653/v1/2024.wmt-1.2}.
\newblock URL \url{https://aclanthology.org/2024.wmt-1.2/}.

\bibitem[{Gemini Team}(2025)]{comanici2025gemini25pushingfrontier}
{Gemini Team}.
\newblock Gemini 2.5: Pushing the frontier with advanced reasoning, multimodality, long context, and next generation agentic capabilities, 2025.
\newblock URL \url{https://arxiv.org/abs/2507.06261}.

\bibitem[Guerreiro et~al.(2024)Guerreiro, Rei, Stigt, Coheur, Colombo, and Martins]{guerreiro-etal-2024-xcomet}
Guerreiro, N.~M., Rei, R., Stigt, D.~v., Coheur, L., Colombo, P., and Martins, A. F.~T.
\newblock x{COMET}: Transparent machine translation evaluation through fine-grained error detection.
\newblock \emph{Transactions of the Association for Computational Linguistics}, 12:\penalty0 979--995, 2024.
\newblock \doi{10.1162/tacl_a_00683}.
\newblock URL \url{https://aclanthology.org/2024.tacl-1.54/}.

\bibitem[Huang et~al.(2025)Huang, Ding, Pan, and Zhang]{huang2025englishcentrictrainingreinforcementlearning}
Huang, S., Ding, Y., Pan, J., and Zhang, Y.
\newblock Beyond english-centric training: How reinforcement learning improves cross-lingual reasoning in llms, 2025.
\newblock URL \url{https://arxiv.org/abs/2509.23657}.

\bibitem[Jiang et~al.(2025)Jiang, Wang, Yan, Liu, Zhu, Zhang, and Gao]{jiang-etal-2025-large-language}
Jiang, J., Wang, J., Yan, Y., Liu, Y., Zhu, J., Zhang, M., and Gao, L.
\newblock Do large language models excel in complex logical reasoning with formal language?
\newblock In Christodoulopoulos, C., Chakraborty, T., Rose, C., and Peng, V. (eds.), \emph{Proceedings of the 2025 Conference on Empirical Methods in Natural Language Processing}, pp.\  16889--16914, Suzhou, China, November 2025. Association for Computational Linguistics.
\newblock ISBN 979-8-89176-332-6.
\newblock \doi{10.18653/v1/2025.emnlp-main.855}.
\newblock URL \url{https://aclanthology.org/2025.emnlp-main.855/}.

\bibitem[Juraska et~al.(2024)Juraska, Deutsch, Finkelstein, and Freitag]{juraska-etal-2024-metricx}
Juraska, J., Deutsch, D., Finkelstein, M., and Freitag, M.
\newblock {M}etric{X}-24: The {G}oogle submission to the {WMT} 2024 metrics shared task.
\newblock In Haddow, B., Kocmi, T., Koehn, P., and Monz, C. (eds.), \emph{Proceedings of the Ninth Conference on Machine Translation}, pp.\  492--504, Miami, Florida, USA, November 2024. Association for Computational Linguistics.
\newblock \doi{10.18653/v1/2024.wmt-1.35}.
\newblock URL \url{https://aclanthology.org/2024.wmt-1.35/}.

\bibitem[{Kimi Team}(2025)]{kimiteam2025kimik2openagentic}
{Kimi Team}.
\newblock Kimi k2: Open agentic intelligence, 2025.
\newblock URL \url{https://arxiv.org/abs/2507.20534}.

\bibitem[Kocmi et~al.(2024)Kocmi, Zouhar, Federmann, and Post]{kocmi-etal-2024-navigating}
Kocmi, T., Zouhar, V., Federmann, C., and Post, M.
\newblock Navigating the metrics maze: Reconciling score magnitudes and accuracies.
\newblock In Ku, L.-W., Martins, A., and Srikumar, V. (eds.), \emph{Proceedings of the 62nd Annual Meeting of the Association for Computational Linguistics (Volume 1: Long Papers)}, pp.\  1999--2014, Bangkok, Thailand, August 2024. Association for Computational Linguistics.
\newblock \doi{10.18653/v1/2024.acl-long.110}.
\newblock URL \url{https://aclanthology.org/2024.acl-long.110/}.

\bibitem[Kocmi et~al.(2025)Kocmi, Arkhangorodsky, Berard, Blunsom, Cahyawijaya, Dehaze, Fadaee, Frosst, Galle, Gomez, Govindarajan, Ko, Kreutzer, Marchisio, {\"U}st{\"u}n, Vincent, and Zhang]{kocmi-etal-2025-command}
Kocmi, T., Arkhangorodsky, A., Berard, A., Blunsom, P., Cahyawijaya, S., Dehaze, T., Fadaee, M., Frosst, N., Galle, M., Gomez, A., Govindarajan, N., Ko, W.-Y., Kreutzer, J., Marchisio, K., {\"U}st{\"u}n, A., Vincent, S., and Zhang, I.
\newblock Command-a-translate: Raising the bar of machine translation with difficulty filtering.
\newblock In Haddow, B., Kocmi, T., Koehn, P., and Monz, C. (eds.), \emph{Proceedings of the Tenth Conference on Machine Translation}, pp.\  789--799, Suzhou, China, November 2025. Association for Computational Linguistics.
\newblock ISBN 979-8-89176-341-8.
\newblock \doi{10.18653/v1/2025.wmt-1.55}.
\newblock URL \url{https://aclanthology.org/2025.wmt-1.55/}.

\bibitem[Li et~al.(2025)Li, Emad, Padthe, Lanchantin, Yuan, Nguyen, Weston, Li, Wang, Kulikov, and Li]{li2025naturalthoughtsselectingdistillingreasoning}
Li, Y., Emad, Y., Padthe, K., Lanchantin, J., Yuan, W., Nguyen, T., Weston, J., Li, S.-W., Wang, D., Kulikov, I., and Li, X.
\newblock Naturalthoughts: Selecting and distilling reasoning traces for general reasoning tasks, 2025.
\newblock URL \url{https://arxiv.org/abs/2507.01921}.

\bibitem[Li et~al.(2026)Li, Ji, and Tiedemann]{li2026testtimescalingreasoningmodels}
Li, Z., Ji, S., and Tiedemann, J.
\newblock Test-time scaling of reasoning models for machine translation, 2026.
\newblock URL \url{https://arxiv.org/abs/2510.06471}.

\bibitem[Liao et~al.(2025)Liao, Chen, Rajaee, Xu, Herold, Søgaard, de~Rijke, and Monz]{liao2025lostbeginningreasoning}
Liao, B., Chen, X., Rajaee, S., Xu, Y., Herold, C., Søgaard, A., de~Rijke, M., and Monz, C.
\newblock Lost at the beginning of reasoning, 2025.
\newblock URL \url{https://arxiv.org/abs/2506.22058}.

\bibitem[Liu et~al.(2024)Liu, Lin, Hewitt, Paranjape, Bevilacqua, Petroni, and Liang]{liu-etal-2024-lost}
Liu, N.~F., Lin, K., Hewitt, J., Paranjape, A., Bevilacqua, M., Petroni, F., and Liang, P.
\newblock Lost in the middle: How language models use long contexts.
\newblock \emph{Transactions of the Association for Computational Linguistics}, 12:\penalty0 157--173, 2024.
\newblock \doi{10.1162/tacl_a_00638}.
\newblock URL \url{https://aclanthology.org/2024.tacl-1.9/}.

\bibitem[Liu et~al.(2025)Liu, Lyu, Wu, Wang, Luo, Zhang, and Shang]{liu2025newtrendsmodernmachine}
Liu, S., Lyu, C., Wu, M., Wang, L., Luo, W., Zhang, K., and Shang, Z.
\newblock New trends for modern machine translation with large reasoning models, 2025.
\newblock URL \url{https://arxiv.org/abs/2503.10351}.

\bibitem[Lommel et~al.(2013)Lommel, Burchardt, and Uszkoreit]{burchardt-2013-multidimensional}
Lommel, A.~R., Burchardt, A., and Uszkoreit, H.
\newblock Multidimensional quality metrics: a flexible system for assessing translation quality.
\newblock In \emph{Proceedings of Translating and the Computer 35}, London, UK, November 28-29 2013. Aslib.
\newblock URL \url{https://aclanthology.org/2013.tc-1.6/}.

\bibitem[Marjanovic et~al.(2026)Marjanovic, Patel, Adlakha, Aghajohari, BehnamGhader, Bhatia, Khandelwal, Kraft, Krojer, L{\`u}, Meade, Shin, Kazemnejad, Kamath, Mosbach, Stanczak, and Reddy]{marjanovic2026deepseekr}
Marjanovic, S.~V., Patel, A., Adlakha, V., Aghajohari, M., BehnamGhader, P., Bhatia, M., Khandelwal, A., Kraft, A., Krojer, B., L{\`u}, X.~H., Meade, N., Shin, D., Kazemnejad, A., Kamath, G., Mosbach, M., Stanczak, K., and Reddy, S.
\newblock Deepseek-r1 thoughtology: Let{\textquoteright}s think about {LLM} reasoning.
\newblock \emph{Transactions on Machine Learning Research}, 2026.
\newblock ISSN 2835-8856.
\newblock URL \url{https://openreview.net/forum?id=BZwKsiRnJI}.

\bibitem[Min et~al.(2024)Min, Ding, Buratti, Pujar, Kaiser, Jana, and Ray]{min2024beyond}
Min, M.~J., Ding, Y., Buratti, L., Pujar, S., Kaiser, G., Jana, S., and Ray, B.
\newblock Beyond accuracy: Evaluating self-consistency of code large language models with identitychain.
\newblock In \emph{The Twelfth International Conference on Learning Representations}, 2024.
\newblock URL \url{https://openreview.net/forum?id=caW7LdAALh}.

\bibitem[Mistral-AI(2025)]{mistralai2025magistral}
Mistral-AI.
\newblock Magistral, 2025.
\newblock URL \url{https://arxiv.org/abs/2506.10910}.

\bibitem[Nguyen \& Xu(2025)Nguyen and Xu]{nguyen-xu-2025-reasoning}
Nguyen, L. and Xu, Y.
\newblock Reasoning for translation: Comparative analysis of chain-of-thought and tree-of-thought prompting for {LLM} translation.
\newblock In Zhao, J., Wang, M., and Liu, Z. (eds.), \emph{Proceedings of the 63rd Annual Meeting of the Association for Computational Linguistics (Volume 4: Student Research Workshop)}, pp.\  259--275, Vienna, Austria, July 2025. Association for Computational Linguistics.
\newblock ISBN 979-8-89176-254-1.
\newblock \doi{10.18653/v1/2025.acl-srw.17}.
\newblock URL \url{https://aclanthology.org/2025.acl-srw.17/}.

\bibitem[{NVIDIA}(2025)]{nvidia2025nvidianemotronnano2}
{NVIDIA}.
\newblock Nvidia nemotron nano 2: An accurate and efficient hybrid mamba-transformer reasoning model, 2025.
\newblock URL \url{https://arxiv.org/abs/2508.14444}.

\bibitem[Pham et~al.(2025)Pham, Nguyen, Zunjare, Chen, Tseng, and Vu]{pham2025sealqaraisingbarreasoning}
Pham, T., Nguyen, N., Zunjare, P., Chen, W., Tseng, Y.-M., and Vu, T.
\newblock Sealqa: Raising the bar for reasoning in search-augmented language models, 2025.
\newblock URL \url{https://arxiv.org/abs/2506.01062}.

\bibitem[Proietti et~al.(2025)Proietti, Perrella, Zouhar, Navigli, and Kocmi]{proietti-etal-2025-estimating}
Proietti, L., Perrella, S., Zouhar, V., Navigli, R., and Kocmi, T.
\newblock Estimating machine translation difficulty.
\newblock In Christodoulopoulos, C., Chakraborty, T., Rose, C., and Peng, V. (eds.), \emph{Findings of the Association for Computational Linguistics: EMNLP 2025}, pp.\  24261--24285, Suzhou, China, November 2025. Association for Computational Linguistics.
\newblock ISBN 979-8-89176-335-7.
\newblock \doi{10.18653/v1/2025.findings-emnlp.1317}.
\newblock URL \url{https://aclanthology.org/2025.findings-emnlp.1317/}.

\bibitem[{Qwen Team}(2025)]{qwen3technicalreport}
{Qwen Team}.
\newblock Qwen3 technical report, 2025.
\newblock URL \url{https://arxiv.org/abs/2505.09388}.

\bibitem[Son et~al.(2026)Son, Yang, Patel, Agarwal, Ko, Lim, Panda, Kim, Drolia, Choi, Lee, and Yu]{son2026pushingmultilingualreasoningmodels}
Son, G., Yang, D., Patel, H.~L., Agarwal, A., Ko, H., Lim, C., Panda, S., Kim, M., Drolia, N., Choi, D., Lee, K.-H., and Yu, Y.
\newblock Pushing on multilingual reasoning models with language-mixed chain-of-thought, 2026.
\newblock URL \url{https://arxiv.org/abs/2510.04230}.

\bibitem[Sprague et~al.(2025)Sprague, Yin, Rodriguez, Jiang, Wadhwa, Singhal, Zhao, Ye, Mahowald, and Durrett]{sprague2025to}
Sprague, Z.~R., Yin, F., Rodriguez, J.~D., Jiang, D., Wadhwa, M., Singhal, P., Zhao, X., Ye, X., Mahowald, K., and Durrett, G.
\newblock To cot or not to cot? chain-of-thought helps mainly on math and symbolic reasoning.
\newblock In \emph{The Thirteenth International Conference on Learning Representations}, 2025.
\newblock URL \url{https://openreview.net/forum?id=w6nlcS8Kkn}.

\bibitem[Sun et~al.(2024)Sun, Yin, Li, Wu, Qiu, and Kong]{sun2024corex}
Sun, Q., Yin, Z., Li, X., Wu, Z., Qiu, X., and Kong, L.
\newblock Corex: Pushing the boundaries of complex reasoning through multi-model collaboration.
\newblock In \emph{First Conference on Language Modeling}, 2024.
\newblock URL \url{https://openreview.net/forum?id=7BCmIWVT0V}.

\bibitem[Wang et~al.(2025{\natexlab{a}})Wang, Liu, DAI, Zeng, Zhang, Wu, Luo, Li, Tang, He, and Wang]{wang2025agenttts}
Wang, F., Liu, H., DAI, Z., Zeng, J., Zhang, Z., Wu, Z., Luo, C., Li, Z., Tang, X., He, Q., and Wang, S.
\newblock Agent{TTS}: Large language model agent for test-time compute-optimal scaling strategy in complex tasks.
\newblock In \emph{The Thirty-ninth Annual Conference on Neural Information Processing Systems}, 2025{\natexlab{a}}.
\newblock URL \url{https://openreview.net/forum?id=BuYtcTUMyA}.

\bibitem[Wang et~al.(2025{\natexlab{b}})Wang, Meng, Liang, and Zhou]{wang-etal-2025-drt}
Wang, J., Meng, F., Liang, Y., and Zhou, J.
\newblock {DRT}: Deep reasoning translation via long chain-of-thought.
\newblock In Che, W., Nabende, J., Shutova, E., and Pilehvar, M.~T. (eds.), \emph{Findings of the Association for Computational Linguistics: ACL 2025}, pp.\  6770--6782, Vienna, Austria, July 2025{\natexlab{b}}. Association for Computational Linguistics.
\newblock ISBN 979-8-89176-256-5.
\newblock \doi{10.18653/v1/2025.findings-acl.351}.
\newblock URL \url{https://aclanthology.org/2025.findings-acl.351/}.

\bibitem[Wang et~al.(2025{\natexlab{c}})Wang, Zhang, Huang, Yang, Zhang, Huang, and Wang]{wang2025samplingefficient}
Wang, Y., Zhang, P., Huang, S., Yang, B., Zhang, Z., Huang, F., and Wang, R.
\newblock Sampling-efficient test-time scaling: Self-estimating the best-of-n sampling in early decoding.
\newblock In \emph{The Thirty-ninth Annual Conference on Neural Information Processing Systems}, 2025{\natexlab{c}}.
\newblock URL \url{https://openreview.net/forum?id=BcKYVmh3yH}.

\bibitem[Wang et~al.(2025{\natexlab{d}})Wang, Zhang, Qin, Zhi, Li, Huang, Li, and Deng]{wang-etal-2025-exploracoder}
Wang, Y., Zhang, Y., Qin, Z., Zhi, C., Li, B., Huang, F., Li, Y., and Deng, S.
\newblock {E}xplora{C}oder: Advancing code generation for multiple unseen {API}s via planning and chained exploration.
\newblock In Che, W., Nabende, J., Shutova, E., and Pilehvar, M.~T. (eds.), \emph{Proceedings of the 63rd Annual Meeting of the Association for Computational Linguistics (Volume 1: Long Papers)}, pp.\  18124--18145, Vienna, Austria, July 2025{\natexlab{d}}. Association for Computational Linguistics.
\newblock ISBN 979-8-89176-251-0.
\newblock \doi{10.18653/v1/2025.acl-long.887}.
\newblock URL \url{https://aclanthology.org/2025.acl-long.887/}.

\bibitem[Wei et~al.(2022)Wei, Wang, Schuurmans, Bosma, Ichter, Xia, Chi, Le, and Zhou]{wei-cot}
Wei, J., Wang, X., Schuurmans, D., Bosma, M., Ichter, B., Xia, F., Chi, E.~H., Le, Q.~V., and Zhou, D.
\newblock Chain-of-thought prompting elicits reasoning in large language models.
\newblock In \emph{Proceedings of the 36th International Conference on Neural Information Processing Systems}, NIPS '22, Red Hook, NY, USA, 2022. Curran Associates Inc.
\newblock ISBN 9781713871088.

\bibitem[Wu et~al.(2025{\natexlab{a}})Wu, Aycock, and Monz]{wu-etal-2025-please}
Wu, D., Aycock, S., and Monz, C.
\newblock Please translate again: Two simple experiments on whether human-like reasoning helps translation.
\newblock In Christodoulopoulos, C., Chakraborty, T., Rose, C., and Peng, V. (eds.), \emph{Proceedings of the 2025 Conference on Empirical Methods in Natural Language Processing}, pp.\  20435--20451, Suzhou, China, November 2025{\natexlab{a}}. Association for Computational Linguistics.
\newblock ISBN 979-8-89176-332-6.
\newblock \doi{10.18653/v1/2025.emnlp-main.1031}.
\newblock URL \url{https://aclanthology.org/2025.emnlp-main.1031/}.

\bibitem[Wu et~al.(2025{\natexlab{b}})Wu, Xu, Yuan, Haffari, Wan, Luo, and Zhang]{10.1162/TACL.a.25}
Wu, M., Xu, J., Yuan, Y., Haffari, G., Wan, L., Luo, W., and Zhang, K.
\newblock (perhaps) beyond human translation: Harnessing multi-agent collaboration for translating ultra-long literary texts.
\newblock \emph{Transactions of the Association for Computational Linguistics}, 13:\penalty0 901--922, 07 2025{\natexlab{b}}.
\newblock ISSN 2307-387X.
\newblock \doi{10.1162/TACL.a.25}.
\newblock URL \url{https://doi.org/10.1162/TACL.a.25}.

\bibitem[Yu et~al.(2025)Yu, Zhang, Zhang, Liang, Zhang, Zhang, Khademi, Awadalla, Wang, Yang, and Wei]{yu-etal-2025-chain}
Yu, Y., Zhang, Y., Zhang, D., Liang, X., Zhang, H., Zhang, X., Khademi, M., Awadalla, H.~H., Wang, J., Yang, Y., and Wei, F.
\newblock Chain-of-reasoning: Towards unified mathematical reasoning in large language models via a multi-paradigm perspective.
\newblock In Che, W., Nabende, J., Shutova, E., and Pilehvar, M.~T. (eds.), \emph{Proceedings of the 63rd Annual Meeting of the Association for Computational Linguistics (Volume 1: Long Papers)}, pp.\  24914--24937, Vienna, Austria, July 2025. Association for Computational Linguistics.
\newblock ISBN 979-8-89176-251-0.
\newblock \doi{10.18653/v1/2025.acl-long.1213}.
\newblock URL \url{https://aclanthology.org/2025.acl-long.1213/}.

\bibitem[Zebaze et~al.(2025)Zebaze, Bawden, and Sagot]{zebaze2025llmreasoningmachinetranslation}
Zebaze, A., Bawden, R., and Sagot, B.
\newblock Llm reasoning for machine translation: Synthetic data generation over thinking tokens, 2025.
\newblock URL \url{https://arxiv.org/abs/2510.11919}.

\bibitem[Zhang et~al.(2024)Zhang, Zhang, Li, and Xing]{zhang-etal-2024-evaluating}
Zhang, Y., Zhang, H., Li, L., and Xing, E.
\newblock Evaluating step-by-step reasoning through symbolic verification.
\newblock In Duh, K., Gomez, H., and Bethard, S. (eds.), \emph{Findings of the Association for Computational Linguistics: NAACL 2024}, pp.\  2984--3002, Mexico City, Mexico, June 2024. Association for Computational Linguistics.
\newblock \doi{10.18653/v1/2024.findings-naacl.188}.
\newblock URL \url{https://aclanthology.org/2024.findings-naacl.188/}.

\end{thebibliography}
\bibliographystyle{icml2025}

%%%%%%%%%%%%%%%%%%%%%%%%%%%%%%%%%%%%%%%%%%%%%%%%%%%%%%%%%%%%%%%%%%%%%%%%%%%%%%%
%%%%%%%%%%%%%%%%%%%%%%%%%%%%%%%%%%%%%%%%%%%%%%%%%%%%%%%%%%%%%%%%%%%%%%%%%%%%%%%
% APPENDIX
%%%%%%%%%%%%%%%%%%%%%%%%%%%%%%%%%%%%%%%%%%%%%%%%%%%%%%%%%%%%%%%%%%%%%%%%%%%%%%%
%%%%%%%%%%%%%%%%%%%%%%%%%%%%%%%%%%%%%%%%%%%%%%%%%%%%%%%%%%%%%%%%%%%%%%%%%%%%%%%
\newpage
\appendix
\onecolumn

%%%%%%%%%%%%%%%%%%%%%%%%%%%%%%%%%%%%%%%%%%%%%%%%%%%%%%%%%%%%%%%%%%%%%%%%%%%%%%%
%%%%%%%%%%%%%%%%%%%%%%%%%%%%%%%%%%%%%%%%%%%%%%%%%%%%%%%%%%%%%%%%%%%%%%%%%%%%%%%
\section{Experimental Setups}
\label{app:experimental-setups}
In this paper, we consider nine language pairs to study the reasoning capabilities of RLMs using WMT24++ benchmark, including Arabic (ar), Czech (cs), Farsi (fa), French (fr), Hindi (hi), Japanese (ja), Korean (ko), Russian (ru), and Chinese (zh). In selecting language pairs, we made sure to cover a diverse set of languages. 
\subsection{Evaluation prompt}
We employ the originally suggested WMT24++ evaluation prompt with a
 slight adjustment~\cite{deutsch-etal-2025-wmt24}, presented in the following box. 
 \begin{tcolorbox}[
  colback=white,
  colframe=black,
  title=WMT24++ Evaluation Prompt,
  fonttitle=\bfseries,
]
You are a professional ${src\_lang}$ to ${tgt\_lang}$ translator, tasked with providing translations suitable for use in ${tgt\_region}$ (${tgt\_code}$). Your goal is to accurately convey the meaning and nuances of the original ${src\_lang}$ text while adhering to ${tgt\_lang}$ grammar, vocabulary, and cultural sensitivities. Produce only the ${tgt\_lang}$ translation, without any additional explanations or commentary. Please translate the following ${src\_lang}$ text into ${tgt\_lang}$ (${tgt\_code}$):
\end{tcolorbox}
\subsection{Training data distribution}

\noindent
\begin{minipage}{0.99\linewidth}
\end{minipage}
\hfill
\begin{minipage}{0.99\linewidth}
\centering
\includegraphics[width=\linewidth]{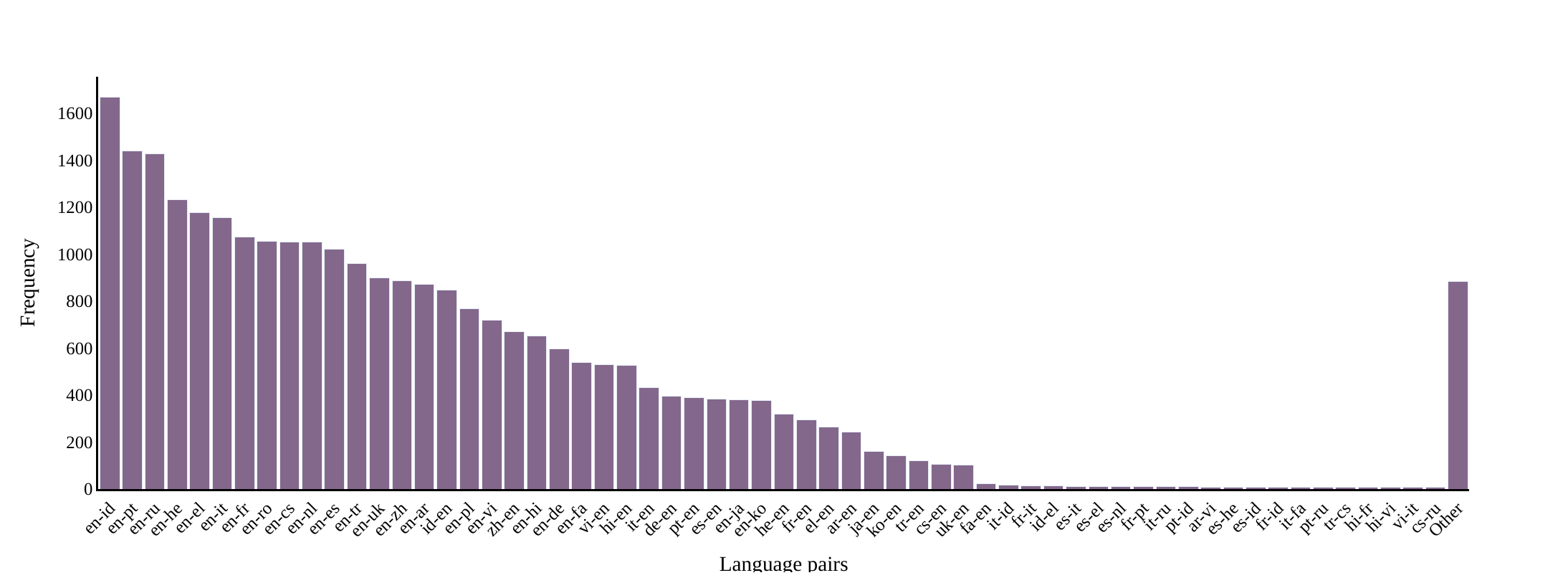}
\captionof{figure}{The distribution of language pairs in 28k structured reasoning traces as the training data. Most language pairs have a very low frequency. We put them in the \textit{Other} category for presentation.}
\label{fig:lang-dist}
\end{minipage}

\newpage
\section{A Reasoning Example}
\label{app:reasoning-example}
A generated reasoning trace and final translation for the given en-fr example by \deepseek{}.
\begin{tcolorbox}[
  colback=white,
  colframe=black,
  boxrule=0.4pt,
  arc=1mm,
  left=6pt,
  right=6pt,
  top=6pt,
  bottom=6pt
]

\textbf{Source Text}  
\textcolor{sectiongray}{(Input)}  
\vspace{2pt}

While the Louisiana Republican can avoid confronting the issue for now, on January 19 the country will face its deadline to pass four of 12 annual spending bills. Johnson has publicly stated that he would not support a "continuing resolution" (CR) to extend the deadline on these four bills to the February 2 due date for the remaining eight.

\medskip
\hrule
\medskip

\textbf{Reasoning}   
\vspace{2pt}

Hmm, the user wants me to translate an English political text about US budget deadlines into French. They've specified I should act as a professional translator focusing on French (France) conventions without any explanations.

This text discusses Louisiana Republican Mike Johnson's position on government spending bills. The key elements are: temporary avoidance of an issue, January 19 deadline for four bills, Johnson's rejection of a "continuing resolution" (CR), and the February 2 deadline for eight other bills.

For the translation, I need to consider several aspects. "Louisiana Republican" should become "le républicain de Louisiane" to maintain political context. "Confronting the issue" translates best to "confronter ce problème" in French political discourse. The term "continuing resolution" requires keeping the English abbreviation "CR" with French explanation in parentheses - "une 'continuing resolution' (CR)" - as this is standard practice for technical terms.

The dates need European formatting: "19 janvier" and "2 février". For "publicly stated", "a publiquement déclaré" sounds more natural than literal translations. The structure about extending deadlines to February 2 requires careful handling - "reporter l'échéance au 2 février" flows better than word-for-word translation.

I should ensure all verbs agree properly in tense and that the sentence structure maintains the original's urgency while sounding natural in French. The final translation needs to be polished and ready for professional use without any translator's notes.

\medskip
\hrule
\medskip

\textbf{Generated Translation}  
\textcolor{sectiongray}{(Final output)}  
\vspace{2pt}

Si le républicain de Louisiane peut éviter de confronter ce problème pour l'instant, le pays devra faire face, le 19 janvier, à l'échéance pour l'adoption de quatre des douze projets de loi de dépenses annuelles. Johnson a publiquement déclaré qu'il ne soutiendrait pas une "continuing resolution" (CR) pour reporter l'échéance de ces quatre textes au 2 février, date limite pour les huit autres.
\end{tcolorbox}
\newpage
\section{Translation Prompts}
\label{app:translation-prompts}
We use the following prompts to generate translations at each step of our 4-step template using \cmdatranslate{}.
\begin{tcolorbox}[
  enhanced,
  colback=white,
  colframe=FrameGray,
  boxrule=0.5pt,
  arc=1mm,
  left=7pt,right=7pt,top=6pt,bottom=7pt,
  fontupper=\ttfamily\small,
  title=\textbf{Step-wise Translation Prompt Templates},
  coltitle=HeaderFG,
  colbacktitle=HeaderBG,
  boxed title style={
    sharp corners,
    boxrule=0pt,
    left=6pt,right=6pt,top=3pt,bottom=3pt
  },
]

\textbf{Draft (Initial Translation)}\\
Your goal is to translate a piece of text into \{target\_language\} that accurately conveys the meaning and structure of the source text. Your translation should closely reflect the original content without omitting or adding information. If certain contextual details are missing, aim for a general translation that remains flexible across different contexts, and do not include any explanations or commentary. Keep the same number of lines as in the source text.\\
Provide your best single translation of the original text enclosed in triple backticks:
\begin{verbatim}
```{source_text}```
\end{verbatim}
\vspace{3pt}
\hrule
\vspace{3pt}

\textbf{Refinement (Adequacy-Focused Translation)}\\
Now, produce a new translation into \{target\_language\} that focuses primarily on the adequacy of translation.\\

\hrule
\vspace{3pt}

\textbf{Refinement (Fluency-Focused Translation)}\\
Now, produce a new translation into \{target\_language\} that focuses primarily on the fluency of translation, ensuring it reads as if it were originally written in \{target\_language\}. Provide only one refined translation enclosed in triple backticks and do not output anything else after that.\\

\hrule
\vspace{3pt}

\textbf{Proofreading (Final Translation)}\\
You are tasked with proofreading and final editing of a translation. You will be provided with three different translations: first draft, translation focused on adequacy, and translation focused on fluency. Your goal is to provide a polished final translation of the source text into \{target\_language\}. Keep the same number of lines as in the source text.

For your reference, below are the source text, the draft translation, and refined translations, each enclosed in triple backticks:

\begin{verbatim}
Source ({source_language}):
```{source_text}```

Draft translation:
```{preprocess_analysis}```

Refined translation (adequacy):
```{draft_translation}```

Refined translation (fluency):
```{refined_translation}```
\end{verbatim}

Proofread the refined text for grammar, spelling, punctuation, terminology, and overall fluency. Ensure the translation accurately reflects the original meaning, intent, style, and structure, while sounding fully natural and idiomatic in \{target\_language\}. Provide only the final polished translation enclosed in triple backticks and nothing else.

\end{tcolorbox}
\newpage
\section{Static Template}

\label{app:static-reasoning-template}
Here we bring an example of the static template. At each step, all the document translation is presented, making the reasoning trace lengthy and inefficient.
\begin{tcolorbox}[
  width=\textwidth,
  enhanced,
  colback=white,
  colframe=FrameGray,
  boxrule=0.5pt,
  arc=1mm,
  left=6pt,right=6pt,top=5pt,bottom=6pt,
  fontupper=\ttfamily\small,
  title=\textbf{Static Reasoning Template},
  coltitle=HeaderFG,
  colbacktitle=HeaderBG,
  colframe=FrameGray,
  boxed title style={
    sharp corners,
    boxrule=0pt,
    left=6pt,right=6pt,top=3pt,bottom=3pt
  },
]
\textbf{Step 1 --- Initial Draft}\\
Let me start by producing an initial, faithful translation of the source text into \{target\_language\}, which will serve as the foundation for later refinements.\\
\{initial\_draft\}\\

\vspace{3pt}\hrule\vspace{3pt}

\textbf{Step 2 --- Adequacy}\\
In this step, I focus on adequacy, ensuring that the full meaning of the challenging sentences is conveyed accurately and completely in \{target\_language\}.\\
\{adequacy\_focused\_translation\}\\

\vspace{3pt}\hrule\vspace{3pt}

\textbf{Step 3 --- Fluency}\\
At this stage, I refine the challenging sentences, ensuring they read smoothly and naturally in \{target\_language\}, as if originally written in it.\\
\{fluency\_focused\_translation\}\\

\vspace{3pt}\hrule\vspace{3pt}

\textbf{Step 4 --- Final Translation}\\
Finally, I bring together all the work from the previous steps, addressing any remaining issues or translation errors to produce a polished final version in \{target\_language\}.\\
\{final\_translation\}

\end{tcolorbox}

\newpage
\section{Structured Reasoning Templates}
\label{app:structured-reasoning-template}
To make a diverse training set and avoid overfitting on the predefined sentences in the structured reasoning template, we consider a diverse set of transitional sentences for each step. We randomly pick a sentence for each step to generate a reasoning trace. Below, you find the set of sentences we considered. 

\begin{tcolorbox}[
  width=\textwidth,
  enhanced,
  colback=white,
  colframe=FrameGray,
  boxrule=0.5pt,
  arc=1mm,
  left=7pt,right=7pt,top=6pt,bottom=7pt,
  fontupper=\ttfamily\small,
  title=\textbf{Transitional Sentences for Structured Reasoning},
  coltitle=HeaderFG,
  colbacktitle=HeaderBG,
  boxed title style={
    sharp corners,
    boxrule=0pt,
    left=6pt,right=6pt,top=3pt,bottom=3pt
  },
]

\textbf{Initial Draft}
\begin{itemize}
  \item Let me start by producing an initial, faithful translation of the source text into \{target\_language\}, which will serve as the foundation for later refinements.
  \item To begin, I’ll create an initial, precise translation of the given text into \{target\_language\}, laying the groundwork for further adjustments.
  \item I’ll translate the source text faithfully into \{target\_language\} to form the base for refinement.
  \item For the first step, I’ll produce a faithful initial translation into \{target\_language\}.
  \item Let’s begin with an accurate initial translation of the source into \{target\_language\}, which I’ll refine in later steps if needed.
\end{itemize}

\hrule
\vspace{3pt}

\textbf{Adequacy Refinement}
\begin{itemize}
  \item In this step, I focus on adequacy, ensuring that the full meaning of the challenging sentences is conveyed accurately and completely in \{target\_language\}.
  \item My attention here is on adequacy, making sure the difficult sentences preserve every nuance and idea from the source in \{target\_language\}.
  \item This step centers on adequacy, confirming that each complex sentence accurately communicates the original message in \{target\_language\}.
  \item Adequacy guides my work in this step. I translate the challenging sentences so their sense remains complete and precise in \{target\_language\}.
\end{itemize}

\hrule
\vspace{3pt}

\textbf{Fluency Refinement}
\begin{itemize}
  \item At this stage, I refine the challenging sentences, ensuring they read smoothly and naturally in \{target\_language\}, as if originally written in it.
  \item This step focuses on improving fluency, making the difficult sentences sound idiomatic in \{target\_language\}.
  \item I rework the challenging parts to produce fluent, native-like versions in \{target\_language\}.
  \item I dedicate this step to refining the harder sentences, aiming for smooth and natural expression in \{target\_language\}.
\end{itemize}

\hrule
\vspace{3pt}

\textbf{Final Translation}
\begin{itemize}
  \item Finally, I bring together all the work from the previous steps, addressing any remaining issues to produce a polished final version in \{target\_language\}.
  \item I conclude by integrating all prior revisions and delivering a coherent, high-quality translation in \{target\_language\}.
  \item This step pulls together all prior improvements, correcting residual errors and finalizing the translation in \{target\_language\}.
\end{itemize}

\vspace{4pt}

\end{tcolorbox}
% \caption{Templated prompt used to generate dynamic structured reasoning traces. Curly-braced fields denote placeholders. \textit{One instruction is sampled per step to introduce surface-level diversity while preserving a fixed reasoning structure.}}
\newpage
\section{Direct Translation Baseline}
\label{app:direct-translation-template}
Here we present the templates we use for direct translation baseline. In this setup, we have discarded the reasoning traces, and one of the randomly picked following sentences is put within the thinking tokens.

\begin{tcolorbox}[
  enhanced,
  colback=white,
  colframe=FrameGray,
  boxrule=0.5pt,
  arc=1mm,
  left=7pt,right=7pt,top=6pt,bottom=7pt,
  fontupper=\ttfamily\small,
  title=\textbf{Direct Translation Templates (No Reasoning)},
  coltitle=HeaderFG,
  colbacktitle=HeaderBG,
  boxed title style={
    sharp corners,
    boxrule=0pt,
    left=6pt,right=6pt,top=3pt,bottom=3pt
  },
]

\begin{itemize}
  \item The user is asking for a translation from \{source\_language\} into \{target\_language\}. Analyzing the source text, I noticed that it is not particularly challenging to translate. Therefore, I will skip the reasoning steps and proceed directly with the translation.
  \item This looks like an easy translation request. I’ll skip step-by-step reasoning and directly translate it into \{target\_language\}.
  \item Translate this from \{source\_language\} to \{target\_language\}. Since the text doesn’t appear complex, I’ll skip detailed reasoning and translate it directly.
  \item A translation from \{source\_language\} to \{target\_language\} is requested. The source seems easy, so I’ll provide a direct translation right away.
  \item Let’s translate this into \{target\_language\}. The text isn’t difficult, so I’ll proceed without intermediate reasoning.
  \item I understand the task: translate the given text to \{target\_language\}. It’s straightforward, so I’ll skip the reasoning process.
  \item This is a straightforward translation query. I’ll respond with a direct translation into \{target\_language\}, without a stepwise explanation.
  \item The user needs a translation from \{source\_language\} into \{target\_language\}. I’ll create the final translation directly, as the text poses no challenge.
  \item I’ll translate this from \{source\_language\} into \{target\_language\}. Because the content is easy, I won’t follow a step-by-step translation process.
  \item The task is to translate from \{source\_language\} to \{target\_language\}. The text looks simple enough, so I’ll go straight to the translation.
\end{itemize}

\vspace{4pt}

\end{tcolorbox}

\end{document}